\documentclass[10pt]{article}
\usepackage{csw2008}
\usepackage{graphicx}
\usepackage{times}

\begin{document}

\cswtitleblock{
\cswtitle{Towards A Self-Organized Agent-Based Simulation Model for Exploration of Human Synaptic Connections}
\cswauthor{\"{O}nder G\"{u}rcan}{onder.gurcan@ege.edu.tr}
\cswaddress{Ege University, Computer Engineering Department, Izmir, Turkey\\Paul Sabatier University, IRIT, Institut de Recherche Informatique de Toulouse, France}
\cswauthor{Carole Bernon}{carole.bernon@irit.fr}
\cswaddress{Paul Sabatier University, IRIT, Institut de Recherche Informatique de Toulouse, France}
\cswauthor{Kemal S. T\"{u}rker}{ksturker@ku.edu.tr}
\cswaddress{Koc University, Faculty of Medicine, Istanbul, Turkey}
}

\begin{abstract}
In this paper, the early design of our self-organized agent-based simulation model for exploration of synaptic connections 
that faithfully generates what is observed in natural situation is given. While we take inspiration from neuroscience, our 
intent is not to create a veridical model of processes in neurodevelopmental biology, nor to represent a real biological system. 
Instead, our goal is to design a simulation model that learns acting in the same way of human nervous system by using findings 
on human subjects using reflex methodologies in order to estimate unknown connections.
\end{abstract}

\section{Introduction}
The enormous complexity and the incredible precision of neuronal connectivity have fascinated researchers for a long time.
Although considerable advances have been made during last decades in determining this cellular machinery, understanding how neuronal 
circuits are wired is still one of the holy grails of neuroscience. 
Neuroscientists still rely upon the knowledge
that is obtained in animal studies. Thus, there remains a lack for
human studies revealing functional connectivity at the network level.
This lack might be bridged by novel computational modeling approaches
that learn the dynamics of the networks over time. Such computational
models can be used to put current findings together to obtain the
global picture and to predict hypotheses to lead future experiments.
In this sense, a self-organized agent-based simulation model for exploration of synaptic connectivity is designed that faithfully 
generates what is observed in natural situation. The simulation model uses findings on human subjects using reflex methodologies to 
the computer simulations in order to estimate unknown connections.

Remaining of this paper is organized as follows. Section 2 gives background information, 
section 3 introduces  our simulation model and section 4 summarizes the related work. 
Finally, section 5 gives the future work and concludes the paper.

\section{Background}
Roughly speaking, the central nervous system (CNS) is composed of excitable cells: neurons \& muscles. A typical neuron can be divided into
three functionally distinct parts, dendrites, soma and axon. The dendrites collect synaptic potentials from other neurons and transmits
them to the soma. The soma performs an important non-linear processing step (called \textit{integrate
\& fire model}): If the total synaptic potential exceeds a certain
threshold (makes the neuron membrane potential to \textit{depolarize} to the
threshold), then a spike is generated \cite{Gerstner2002}. A spike is transmitted to another neurons via synapses.
Most synapses occur between an axon terminal of one (presynaptic) neuron and a dendrite 
or the soma of a second (postsynaptic) neuron, or between an axon terminal and a second axon terminal (presynaptic modulation). When 
a spike transmitted by the presynaptic neuron reaches to a synapse, a post-synaptic potential (PSP) occurs on the postsynaptic neuron. 
This PSP can either excite or inhibit a postsynaptic neuron's ability to generate a spike.

To study functional connection of neurons in human
subjects it has been customary to use stimulus-evoked changes in the
discharge probability and rate of one or more \textit{motor units}
in response to stimulation of a set of peripheral afferents or cortico-spinal
fibers. These are the most common ways to investigate the workings
of peripheral and central pathways in human subjects. Although these
are indirect methods of studying human nervous system, they are nevertheless
extremely useful as there is no other method available yet to record
synaptic properties directly in human subjects.
Motor units are composed of one or more alpha-motoneurons and all of the corresponding muscle fibers they innervate. When motor 
units are activated, all of the muscle fibers they innervate contract. The output from the system is through the motoneurons, which 
is measured by reflex recordings from muscle. As output, the instantaneous discharge frequency values against the time of the
stimulus and has recently been used to examine reflex effects on motoneurons,
as well as the sign of the net common input that underlies the synchronous
discharge of human motor units (for a review, see \cite{Turker2005}). However, most 
of the synaptic input to motoneurons from peripheral neurons 
does not go directly to motoneurons, but rather to interneurons 
(whose synaptic connectivity is unknown) that synapse with the motoneurons. 

\section{An Agent-based Simulation Model for Human Motor Units}
For exploring synaptic connectivity in human CNS, we designed and
implemented a self-organized agent-based simulation model. 
Since it seems as a strong candidate for the simulation
work and hence the solution to the problem of putting information
together to predict hypotheses for future studies \cite{Gurcan2010},
we have chosen agent-based modeling and simulation (ABMS) technique.
ABMS is a new approach to modeling systems and is composed of interacting,
autonomous agents \cite{Macal2006}. It is a powerful and flexible
tool for understanding complex adaptive systems such as biological
systems. 

\subsection{Approach to Self-Organization}
Our agent-based simulation model uses the AMAS theory \cite{Capera2003}
to provide agents with adaptive capabilities. This adaptiveness is based on cooperative behavior
which, in this context, means that an agent does all it can to always
help the most annoyed agent (including itself) in the system. When
faced with several problems at the same time, an agent is able to
compute a degree of criticality in order to express how much these
problems are harmful for its own local goal. Considering this criticality,
as well as those of the agents it interacts with, an agent is therefore
able to decide what is the most cooperative action it has to undertake.
The importance of the anomalies and how they are combined emerges
from a cooperative self-adjusting process taking feedbacks into account.

Bernon et al. \cite{Bernon2009} proposed an approach resting on
this theory for engineering self-modeling systems, inwhich same 
type of agents are all designed alike and all agents consist of four behavioral layers.
An agent owns first a \textit{nominal behavior} which represents its behavior when no 
situations that are harmful for its cooperative state are encountered. If a harmful 
situation occurs (such a situation is called a \textit{non-cooperative situation}, or NCS) it has to be avoided or
overcome by every cooperative agent. Therefore, when an agent detects a NCS, at any time during
its lifecycle, it has to adopt a behavior that is able to process
this NCS for coming back to a cooperative state. This provides an
agent with learning capabilities and makes it constantly adapt to
new situations that are judged harmful. The first behavior an agent
tries to adopt to overcome a NCS is a \textit{tuning behavior} in
which it tries to adjust its internal parameters. If this tuning is
impossible because a limit is reached or the agent knows that a worst
situation will occur if it adjusts in a given way, it may propagate
the NCS (or an interpretation of it) to other agents that will try
to help it. If such a behavior of tuning fails, an agent adopts a
\textit{reorganisation behavior} in which it tries to change the way
in which it interacts with others (e.g., by changing a link with another
agent, by creating a new one, by changing the way in which it communicates
with another one and so on). In the same way, for many reasons, this
behavior may fail counteracting the NCS and the last kind of behavior
may be adopted by the agent, the one of \textit{evolution}. In this
last step, an agent may create a new one (e.g., for helping it because
it found nobody else) or may accept to disappear (e.g., it was totally
useless). In these two last levels,
propagation of a problem to other agents is always possible if a local
processing is not achieved.

\begin{figure}
\centering\scalebox{0.165}{\includegraphics{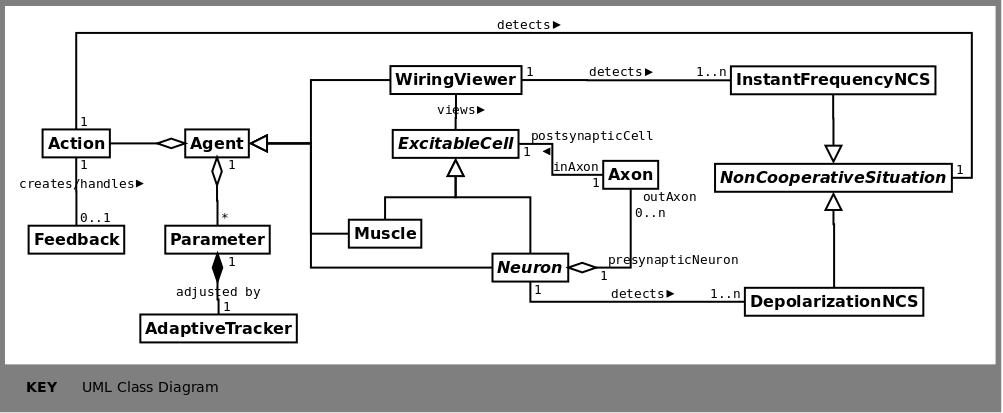}}
\caption{The simulation model for Self-Organizing Agents.} \label{fig:classdiagram}
\end{figure}

\subsection{The Simulation Model}
Figure \ref{fig:classdiagram} shows the conceptual model
of our simulation. \textit{Neuron} and \textit{Muscle} agents are
treated as \textit{ExcitableCell}s. \textit{Axon}s are represented
as connectors between neurons and excitable cells. Unitary behaviors
that an agent is able to do are defined as \textit{Action}s. These
actions can be either for one shot or can be repeated with a specific
interval. Each \textit{Agent} is able to memorize, forget and spontaneously
send feedbacks related to non-desired configuration of inputs (by
detecting NCSs). Each agent has various internal parameters (\textit{Parameter}). When
an agent receives feedbacks from one or more incoming entries, it
is able to adjust its internal parameters or retro-propagates a \textit{Feedback}
to its own entries. For adjusting parameters of agents we used \textit{AdaptiveTracker}s. 
Tuning a parameter for an agent consists
in finding its right value within an interval considering that this
value may evolve with time \cite{Lemouzy2010}. Adaptive trackers
allow this tuning depending on the feedbacks the agent gets from its
environment.

In the AMAS approach, a system is said \textit{functionally adequate}
if it produces the function for which it was conceived, according
to the viewpoint of an external observer who knows its finality. The
external observer in our model is a \textit{WiringViewer} agent. A\textit{
WiringViewer} agent is used to trigger the recruitment of synaptic
connections and the functional connectivity of the neural system.
It monitors and records the outputs of the neural system that take place over
time to compare the simulated (running) data to reference data for detecting NCSs. 
Reference data could be either experimental data or a statistical mean of several
experimental data. the \textit{WiringViewer} agent detects a \textit{InstantFrequencyNCS} 
when an instant frequency of the spike produced by a \textit{Neuron} agent it views is not good.

The nominal behaviour of a \textit{Neuron} agent is to realize \textit{integrate
\& fire model}. As a cooperative behaviour it detects \textit{DepolarizationNCS} 
(the depolarization of a \textit{Neuron} agent can be either lower than needed, higher
than needed or good). Since ``neurons fire together, wire together'', depolarization is crucial for \textit{Neuron} agents. 
After this detection, it sends feedbacks to all its presynaptic agents.  
A \textit{Neuron} agent, receiving either a \textit{DepolarizationNCS} or \textit{InstantFrequencyNCS} feedback, 
tries to increase its PSP or tries to find another \textit{Neuron} agent to help it.

\section{Related Work}
In the literature, there are many models for the self-organization of neuronal
networks. Schoenharl et al. \cite{Schoenharl2005} developed a toolkit for
computational neuroscientists to explore developmental changes in
biological neural networks. However, details of the
methodology used (e.g., how the initial random network is constructed)
and of simulation parameters (e.g., how the threshold parameter for
pruning is obtained) are not clear. Mano et al. \cite{Mano2005} present 
an approach to self-organization in a dynamic neural network by assembling cooperative neuro-agents. 
However, their intent is not to explore synaptic connectivity.
Maniadakis et al. \cite{Maniadakis2009} addresses the development
of brain-inspired models that will be embedded in robotic systems
to support their cognitive abilities. However, this work focuses on 
brain slices rather than reflex pathways
and aims to improve cognitive capabilities of robotic systems rather
than exploring synaptic functional connectivity. 

\section{Conclusion \& Future Work}
Up until now, we have established and implemented a preliminary agent-based simulation model. The next step
will be to enhance and to calibrate the proposed model. We will then compare \textit{in silico} experiments with \textit{in
vitro} biological experiments. As a result of comparison we will either adjust our computatinal model or develop new/improved biological experiments
to revise the biological model. This cycle will proceed until we get satisfactory results.

\bibliographystyle{apalike}
\bibliography{references} 

\begin{thebibliography}{}

\bibitem[Bernon et~al., 2009]{Bernon2009}
Bernon, C., Capera, D., and Mano, J.-P. (2009).
\newblock Engineering self-modeling systems: Application to biology.
\newblock pages 248--263.

\bibitem[Capera et~al., 2003]{Capera2003}
Capera, D., Georg{\'e}, J., Gleizes, M., and Glize, P. (2003).
\newblock The amas theory for complex problem solving based on self-organizing
  cooperative agents.
\newblock In {\em WETICE'03}, page 383, Washington, DC, USA. IEEE Computer
  Society.

\bibitem[Gerstner and Kistler, 2002]{Gerstner2002}
Gerstner, W. and Kistler, W. (2002).
\newblock {\em Spiking Neuron Models}.
\newblock {Cambridge University Press}.

\bibitem[G\"urcan et~al., 2010]{Gurcan2010}
G\"urcan, O., Dikenelli, O., and T\"urker, K.~S. (2010).
\newblock Agent-based exploration of wiring of biological neural networks:
  Position paper.
\newblock In Trumph, R., editor, {\em 20th European Meeting on Cybernetics and
  Systems Research}, pages 509--514.

\bibitem[Lemouzy et~al., 2010]{Lemouzy2010}
Lemouzy, S., Camps, V., and Glize, P. (2010).
\newblock Real time learning of behaviour features for personalised interest
  assessment.
\newblock In {\em Adv. in Practical App. of Agents and Multiagent Systems},
  volume~70 of {\em Adv. in Soft Comp.}, pages 5--14.

\bibitem[Macal and North, 2006]{Macal2006}
Macal, C. and North, M. (2006).
\newblock Tutorial on agent-based modeling and simulation part 2: how to model
  with agents.
\newblock In {\em WSC'06: Proc. of the 38th conf. on Winter simulation}, pages
  73--83.

\bibitem[Maniadakis and Trahanias, 2009]{Maniadakis2009}
Maniadakis, M. and Trahanias, P. (2009).
\newblock Agent-based brain modeling by means of hierarchical cooperative
  coevolution.
\newblock {\em Artificial Life}, 15(3):293--336.

\bibitem[Mano and Glize, 2005]{Mano2005}
Mano, J. and Glize, P. (2005).
\newblock Organization properties of open networks of cooperative neuro-agents.
\newblock In {\em ESANN}, pages 73--78.

\bibitem[Schoenharl, 2005]{Schoenharl2005}
Schoenharl, T. (2005).
\newblock {An Agent Based Approach for the Exploration of Self-Organizing
  Neural Networks}.
\newblock Master's thesis, the Grad. Sch. of the Univ. of Notre Dame.

\bibitem[T{\"u}rker and Powers, 2005]{Turker2005}
T{\"u}rker, K. and Powers, R. (2005).
\newblock Black box revisited: a technique for estimating postsynaptic
  potentials in neurons.
\newblock {\em Trends in neurosciences}, 28(7):379--386.

\end{thebibliography}

\end{document}